\documentclass[5p,,preprint,10pt]{elsarticle}
\makeatletter\if@twocolumn\PassOptionsToPackage{switch}{lineno}\else\fi\makeatother

\usepackage{algorithmic}
\usepackage{algorithm}
\usepackage{tabulary,xcolor}
\usepackage{amsfonts,amsmath,amssymb}
\usepackage[T1]{fontenc}
\usepackage{caption}
\makeatletter
\let\save@ps@pprintTitle\ps@pprintTitle
\def\ps@pprintTitle{\save@ps@pprintTitle\gdef\@oddfoot{\footnotesize\itshape \null\hfill\today}}
\def\hlinewd#1{%
  \noalign{\ifnum0=`}\fi\hrule \@height #1%
  \futurelet\reserved@a\@xhline}

\AtBeginDocument{\ifNAT@numbers \biboptions{sort&compress}\fi}
\makeatother

\usepackage{ifluatex}
\ifluatex
\usepackage{fontspec}
\defaultfontfeatures{Ligatures=TeX}
\usepackage[]{unicode-math}
\unimathsetup{math-style=TeX}
\else 
\usepackage[utf8]{inputenc}
\fi 
\ifluatex\else\usepackage{stmaryrd}\fi

\usepackage{url,multirow,morefloats,floatflt,cancel,tfrupee}
\makeatletter

\AtBeginDocument{\@ifpackageloaded{textcomp}{}{\usepackage{textcomp}}}
\makeatother

\usepackage{color}
\usepackage{hyperref}
\usepackage{colortbl}
\usepackage{xcolor}
\usepackage{cuted}
\usepackage{float}  
\usepackage{caption}
\usepackage{subfigure}
\usepackage{pifont}
\usepackage{amsmath}
\usepackage{multirow}
\usepackage{algorithm,algorithmic}
\usepackage{makecell}
\usepackage{threeparttable}
\usepackage{booktabs}
\usepackage{bm}
\usepackage{diagbox}
\usepackage{graphicx}
\hypersetup{hidelinks,
	colorlinks=true,
	allcolors=blue,
        urlcolor=blue,
	pdfstartview=Fit,
	breaklinks=true}
\usepackage{comment}

\usepackage{colortbl}
\usepackage{xcolor}
\usepackage{cuted}
\usepackage{float}  
\usepackage{caption}
\usepackage{subfigure}
\usepackage{pifont}
\usepackage{amsmath}
\usepackage{makecell}
\usepackage{threeparttable}
\usepackage{booktabs}
\usepackage{bm}
\usepackage{subcaption}
\usepackage{comment}
\usepackage{hyperref}
\hypersetup{hidelinks,
	colorlinks=true,
	allcolors=red,
	pdfstartview=Fit,
	breaklinks=true}
\usepackage[nointegrals]{wasysym}

\makeatletter

\def\mcWidth#1{\csname TY@F#1\endcsname+\tabcolsep}

\def\cAlignHack{\rightskip\@flushglue\leftskip\@flushglue\parindent\z@\parfillskip\z@skip}
\def\rAlignHack{\rightskip\z@skip\leftskip\@flushglue \parindent\z@\parfillskip\z@skip}

\@ifundefined{etal}{}{}

\usepackage{ifxetex}
\ifxetex\else\if@twocolumn\@ifpackageloaded{stfloats}{}{\usepackage{dblfloatfix}}\fi\fi

\AtBeginDocument{
\expandafter\ifx\csname eqalign\endcsname\relax
\def\eqalign#1{\null\vcenter{\def\\{\cr}\openup\jot\m@th
  \ialign{\strut$\displaystyle{##}$\hfil&$\displaystyle{{}##}$\hfil
      \crcr#1\crcr}}\,}
\fi
}

\AtBeginDocument{%
  \@ifpackageloaded{endfloat}%
   {\renewcommand\efloat@iwrite[1]{\immediate\expandafter\protected@write\csname efloat@post#1\endcsname{}}}{\newif\ifefloat@tables}%
}%

\def\BreakURLText#1{\@tfor\brk@tempa:=#1\do{\brk@tempa\hskip0pt}}
\let\lt=<
\let\gt=>
\def\processVert{\ifmmode|\else\textbar\fi}

\@ifundefined{subparagraph}{
\def\subparagraph{\@startsection{paragraph}{5}{2\parindent}{0ex plus 0.1ex minus 0.1ex}%
{0ex}{\normalfont\small\itshape}}%
}{}

\newcommand\role[1]{\unskip}
\newcommand\aucollab[1]{\unskip}
  
\@ifundefined{tsGraphicsScaleX}{\gdef\tsGraphicsScaleX{1}}{}
\@ifundefined{tsGraphicsScaleY}{\gdef\tsGraphicsScaleY{.9}}{}
\def\checkGraphicsWidth{\ifdim\Gin@nat@width>\linewidth
	\tsGraphicsScaleX\linewidth\else\Gin@nat@width\fi}

\def\checkGraphicsHeight{\ifdim\Gin@nat@height>.9\textheight
	\tsGraphicsScaleY\textheight\else\Gin@nat@height\fi}

\def\fixFloatSize#1{}
\let\ts@includegraphics\includegraphics

\def\inlinegraphic[#1]#2{{\edef\@tempa{#1}\edef\baseline@shift{\ifx\@tempa\@empty0\else#1\fi}\edef\tempZ{\the\numexpr(\numexpr(\baseline@shift*\f@size/100))}\protect\raisebox{\tempZ pt}{\ts@includegraphics{#2}}}}

\AtBeginDocument{\def\includegraphics{\@ifnextchar[{\ts@includegraphics}{\ts@includegraphics[width=\checkGraphicsWidth,height=\checkGraphicsHeight,keepaspectratio]}}}

\DeclareMathAlphabet{\mathpzc}{OT1}{pzc}{m}{it}

\def\URL#1#2{\@ifundefined{href}{#2}{\href{#1}{#2}}}

\def\UrlOrds{\do\*\do\-\do\~\do\'\do\"\do\-}%
\g@addto@macro{\UrlBreaks}{\UrlOrds}

\edef\fntEncoding{\f@encoding}

\makeatother

\newif\ifmultipleabstract\multipleabstractfalse%
\usepackage[ singlelinecheck=false ]{caption}

\begin{document}

\begin{frontmatter}

\title{LSTM-CNN: An efficient diagnostic network for Parkinson's disease utilizing dynamic handwriting analysis}

\author[label1]{Xuechao Wang}
\author[label1]{Junqing Huang}
\author[label2]{Sven N{\~omm}}
\author[label1]{Marianna Chatzakou}
\author[label3]{Kadri Medijainen}
\author[label5]{Aaro Toomela} 
\author[label1,label6]{Michael Ruzhansky \corref{cor1}}

\address[label1]{Department of Mathematics: Analysis, Logic and Discrete Mathematics, Ghent University, Ghent, Belgium}
\address[label2]{Department of Software Science, Faculty of Information Technology, Tallinn University of Technology, Akadeemia tee 15 a, 12618, Tallinn, Estonia}
\address[label3]{Institute of Sport Sciences and Physiotherapy, University of Tartu, Puusepa 8, Tartu 51014, Estonia}
\address[label5]{School of Natural Sciences and Health, Tallinn University,
Narva mnt. 25, 10120, Tallinn, Estonia}
\address[label6]{School of Mathematical Sciences, Queen Mary University of London, Mile End Road, London E1 4NS, United Kingdom}

\begin{abstract}
\noindent\textit{Background and objectives:} 
Dynamic handwriting analysis, due to its non-invasive and readily accessible nature, has recently emerged as a vital adjunctive method for the early diagnosis of Parkinson's disease. In this study, we design a compact and efficient network architecture to analyse the distinctive handwriting patterns of patients' dynamic handwriting signals, thereby providing an objective identification for the Parkinson's disease diagnosis. 

\noindent\textit{Methods:} 
The proposed network is based on a hybrid deep learning approach that fully leverages the advantages of both long short-term memory (LSTM) and convolutional neural networks (CNNs). Specifically, the LSTM block is adopted to extract the time-varying features, while the CNN-based block is implemented using one-dimensional convolution for low computational cost. Moreover, the hybrid model architecture is continuously refined under ablation studies for superior performance. Finally, we evaluate the proposed method with its generalization under a five-fold cross-validation, which validates its efficiency and robustness.

\noindent\textit{Results:} 
The proposed network demonstrates its versatility by achieving impressive classification accuracies on both our new DraWritePD dataset ($96.2\%$) and the well-established PaHaW dataset ($90.7\%$). Moreover, the network architecture also stands out for its excellent lightweight design, occupying a mere $0.084$M of parameters, with a total of only $0.59$M floating-point operations. It also exhibits near real-time CPU inference performance, with inference times ranging from $0.106$ to $0.220$s.

\noindent\textit{Conclusions:} 
We present a series of experiments with extensive analysis, which systematically demonstrate the effectiveness and efficiency of the proposed hybrid neural network in extracting distinctive handwriting patterns for precise diagnosis of Parkinson's disease.
\end{abstract}


\end{frontmatter}

\section{Introduction} \label{intro}

Parkinson’s disease (PD) is one of the most widespread and disabling neurodegenerative disorders. The disease is due to the progressive loss of dopaminergic neurons in the middle brain, leading to a decrease in functional, cognitive, and behavioral abilities \cite{hornykiewicz1998biochemical}. Currently, there is no cure, but if PD is diagnosed at an early stage, then the progression of the disease can be significantly slowed with appropriate treatment methods \cite{murman2012early,hauser2010early}. Typical early motor symptoms include resting tremor, rigidity, postural instability, and bradykinesia, i.e., slowness of spontaneous movement \cite{sveinbjornsdottir2016clinical}, especially resting tremor is one of the most common symptoms \cite{hughes1993clinicopathologic}. These motor symptoms, together with non-motor symptoms, have many negative effects on the patient's quality of life, family relationships, and social functioning \cite{dorsey2018global}, while they can also increase the risk of further health complications. This can place a significant economic burden on the individual and society.

\begin{figure*}[htbp]
	\centering  
	\includegraphics[width=\textwidth]{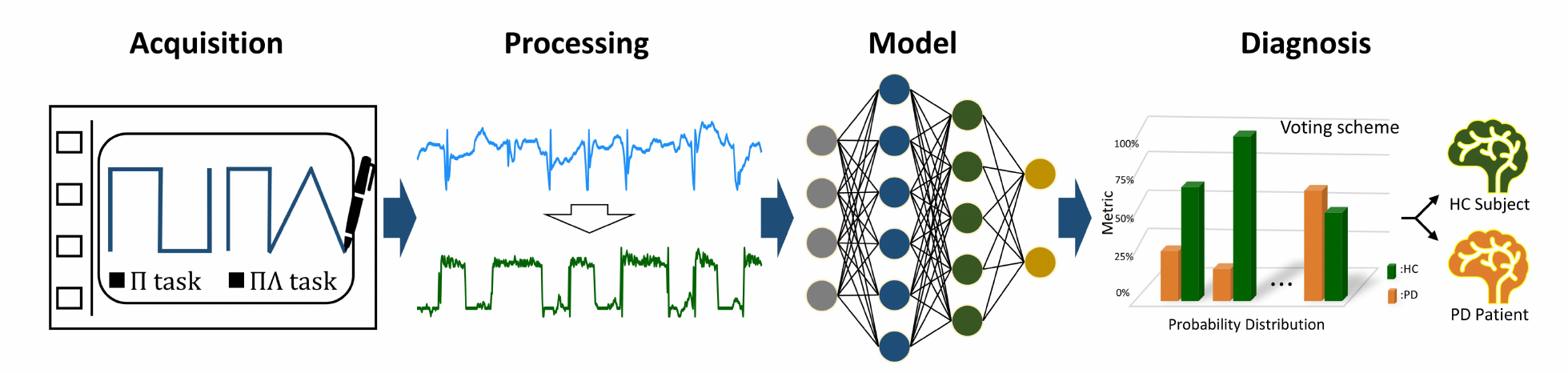}
	\caption{Overview of our proposed framework for Parkinson's disease diagnosis.}
	\label{Fig.1}
\end{figure*}

Handwriting is an extremely common but complex human activity in a variety of leisure and professional settings, which requires fine dexterity skills and involves an intricate blend of cognitive, sensory, and perceptual-motor components \cite{carmeli2003aging}. Changes in it have been well documented to be promising biomarkers for the diagnosis of early PD \cite{thomas2017handwriting, aouraghe2022literature}, and once diagnosed, allow for later neuroprotective interventions. However, related studies have shown that the accuracy of clinical diagnosis is relatively low \cite{schrag2002valid}, while the cost of diagnosis is also quite expensive. Fortunately, a growing body of knowledge provides evidence that it is possible to automatically distinguish between unhealthy and healthy individuals using simple and easy-to-perform handwriting tasks \cite{thomas2017handwriting, aouraghe2023literature}. Therefore, the development of handwriting-based decision support tools is necessary in order to obtain non-invasive and low-cost solutions that can support current standard clinical assessments.

In this research domain, dynamic (online) systems using digital tablets \cite{isenkul2014improved} or Biometric Smart Pens \cite{barth2012combined} can be employed for the diagnosis of PD. Such a device facilitates the capture of various temporal and spatial sequence signals of handwriting, such as pressure and coordinates. Moreover, a critical step in designing such a system is extracting the appropriate features to characterize the unique handwriting patterns of PD patients. To the best of our knowledge, the predominant focus in current methods lies on capturing the global representations of handwriting signals \cite{thomas2017handwriting,aouraghe2023literature}. For instance, handwriting signals undergo feature engineering to extract dynamic features \cite{drotar2013new,drotar2014analysis,drotar2014decision,drotar2016evaluation,valla2022tremor}. Subsequently, to obtain a comprehensive statistical representation of these dynamic features, it is often necessary to calculate single-valued descriptors within the feature vector, such as mean, median, standard deviation, etc. However, compressing the feature vector of arbitrary length into a single value may lead to potential overlook of crucial local details by the diagnostic model. Moreover, other prevalent approaches commonly employ convolutional neural networks (CNNs) for autonomous feature acquisition. Numerous studies \cite{pereira2016deep,pereira2018handwritten,diaz2019dynamically,afonso2019recurrence,nomm2020deep} have effectively addressed the task of extracting features from static two-dimensional ($2$D) images generated from dynamic handwriting signals. However, while these methods may be considered robust alternatives to the artificial engineering, they also only offer a holistic view of the studied handwriting.

To this end, an effective approach for processing sequence signals without losing relevant details involves leveraging the sequence-based neural unit learning paradigm within Recurrent Neural Networks (RNNs) \cite{sherstinsky2020fundamentals}. The online recordings acquired during the writing process can exhibit distinctive time-dependent patterns, which can be effectively employed for distinguishing individuals with PD from healthy controls (HC) \cite{ribeiro2019bag,diaz2021sequence}. Therefore, in this study, unlike compressing feature vectors into single-valued descriptors or reconstructing holistic $2$D images, we investigate the utilization of local one-dimensional ($1$D) dynamic handwriting signals to maximize the preservation of details. Specifically, we propose a compact and efficient hybrid model, LSTM-CNN, that integrates RNNs and CNNs to unveil distinctive handwriting patterns, such as handwriting impairment, among PD patients compared to healthy individuals. This leverages the sequential nature of the data to explicitly incorporate temporal information and provides novel insights into the dynamics handwriting process. 

The main contribution of this study lies in the development of an efficient AI-based framework for PD diagnosis. This framework utilizes a compact hybrid neural network, taking $1$D dynamic signal segments as input. In addition, the designed hybrid model demonstrates outstanding performance within a lightweight structure, achieved through the optimization of network architecture, diagnostic capabilities, and inference efficiency. Beside that, we employ the forward difference algorithm in data processing to extract PD-related derived features, such as resting tremor, from the geometric variables of the handwriting signal. This further enhances the diagnostic performance, while requiring minimal data processing time. Subsequently, we apply a data segmentation technique to enable the proposed hybrid model to focus on the local details while generating sufficient training data. Finally, an inference diagnosis strategy, combined with a majority voting scheme, results in remarkably efficient CPU inference time. 

The paper is organized as follows. Section \ref{sec:related work} introduces the related work on the diagnosis of PD based on dynamic handwriting signals. Section \ref{sec:dataset} provides the reader with the necessary information about the data. The data processing and model details are described in Section \ref{sec:method}. Section \ref{sec:results} presents the main results of the current studies. Finally, the discussion of the results achieved, the limitations of the proposed approaches, as well as the possible future directions are discussed in Section \ref{sec:conlusions}.

\section{Related work}\label{sec:related work}

The application of machine learning techniques in various clinical contexts has gained significant momentum, particularly driving the advancement of automated decision systems for PD diagnosis \cite{thomas2017handwriting, aouraghe2023literature}. We here mainly review the related methods based on dynamic handwriting data and elucidate their interconnections and potential advantages.

A series of outstanding works have been extensively reported, which are grounded in manual feature engineering and aim to extract discriminative features from handwriting signals, subsequently integrating them with traditional machine learning classification models. For example, Drotár et al. \cite{drotar2014analysis} investigate a variety of kinematic-based features, and achieve an $85\%$ accuracy on their publicly available Parkinson’s Disease Handwriting (PaHaW) dataset \cite{drotar2014analysis,drotar2016evaluation}. Furthermore, in \cite{drotar2014decision}, the same author introduce innovative features based on entropy, signal energy and empirical mode decomposition, and enhance diagnostic performance using support vector machine (SVM) classifiers. In a subsequent study \cite{drotar2016evaluation}, Drotár et al. further underscored the significance of novel pressure-related features in the context of PD diagnosis. Besides, Impedovo \cite{impedovo2019velocity} combine classical features with new velocity-based features to extended the handcrafted feature set. This led to improved results on the PaHaW dataset. More recently, Valla et al. \cite{valla2022tremor} explore new derivative-based, angle-type, and integral-like features from the Archimedes spiral graph test, demonstrating their value in dynamic handwriting analysis.

Simultaneously, among well-known contributions based on reconstructing $2$D images, the NewHandPD dataset is introduced in \cite{pereira2016deep}, which is a set of signals extracted from a smart pen, and Pereira et al. convert the dynamic signal into an image and the problem of diagnosing PD is regarded as an image recognition task. This study is one of the first applications of a $2$D deep learning-oriented approach to the diagnosis of PD. Subsequently, the work is further extended in \cite{pereira2018handwritten} and \cite{afonso2019recurrence}. Specifically, Pereira et al. \cite{pereira2018handwritten} combine with various CNN configurations and majority voting schemes to learn texture-oriented features directly from time series-based images. Besides, Afonso et al. \cite{afonso2019recurrence} propose to apply recurrent graph to map signals in the image domain, and feed them to CNNs to learn the appropriate information. Subsequently, the discriminative power of ``dynamically enhanced'' static handwriting images is investigated in \cite{diaz2019dynamically}. The authors propose a static representation that embeds dynamic information and synthesize augmented images using the static and dynamic properties of handwriting. Meanwhile, Nõmm et al. \cite{nomm2020deep} enhance the Archimedes spiral drawing images by controlling the thickness and color of the spiral drawing according to the kinematics and pressure features, all while preserving the original shape of the drawing curve. They achieve an impressive performance using the classic AlexNet \cite{krizhevsky2012imagenet} network.

\begin{figure*}[htbp]	
    \centering
    \subfigure[\textit{$\Pi$} Task and \textit{$\Pi$}\textit{$\Lambda$} Task]
    {
         \begin{minipage}[t]{0.46\linewidth}
            \centering
            \includegraphics[width=0.89\textwidth]{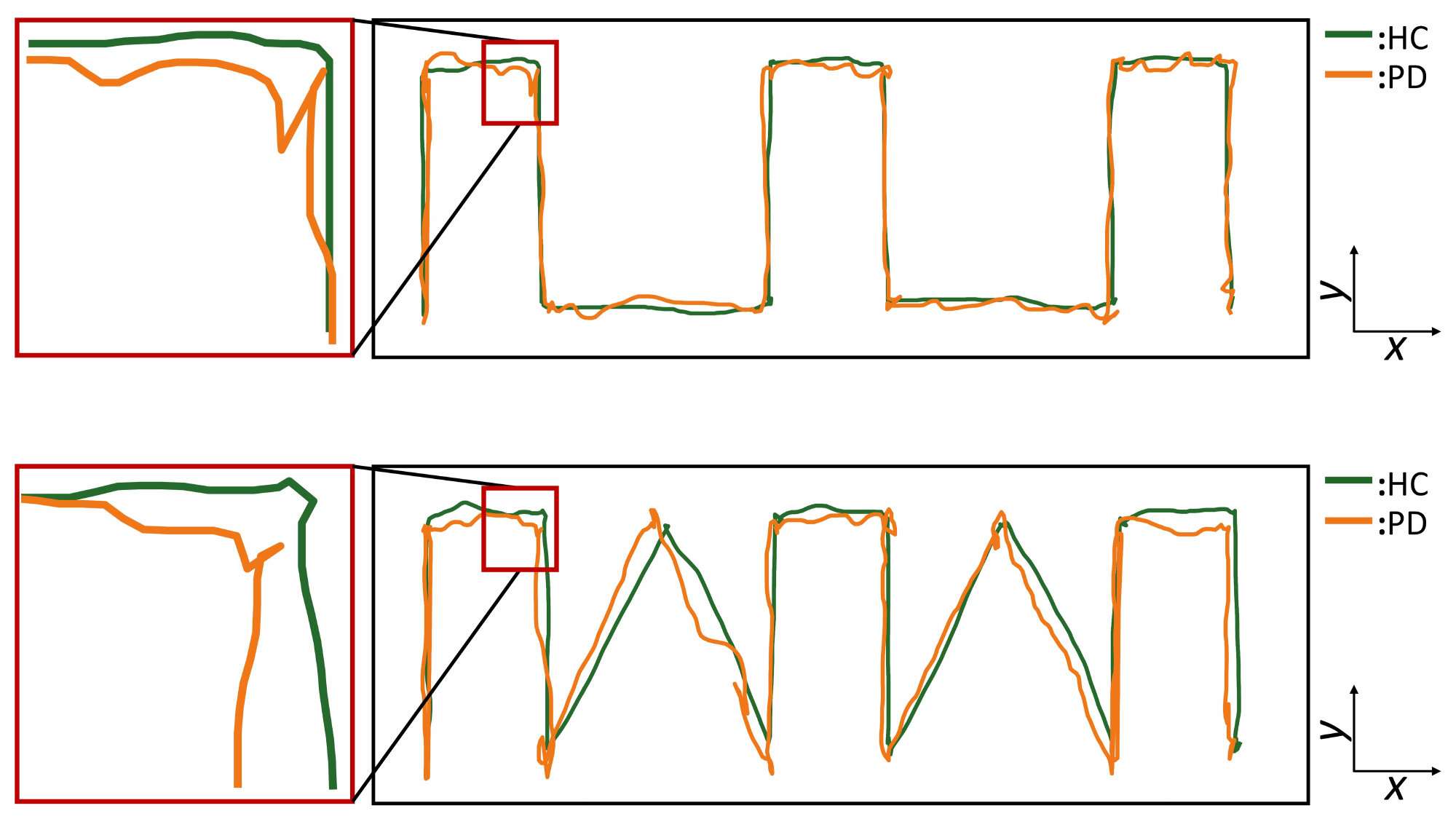}
         \end{minipage}
    }
    \subfigure[Spiral Task] 
    {
        \begin{minipage}[t]{0.46\linewidth}
            \centering
            \includegraphics[width=0.89\textwidth]{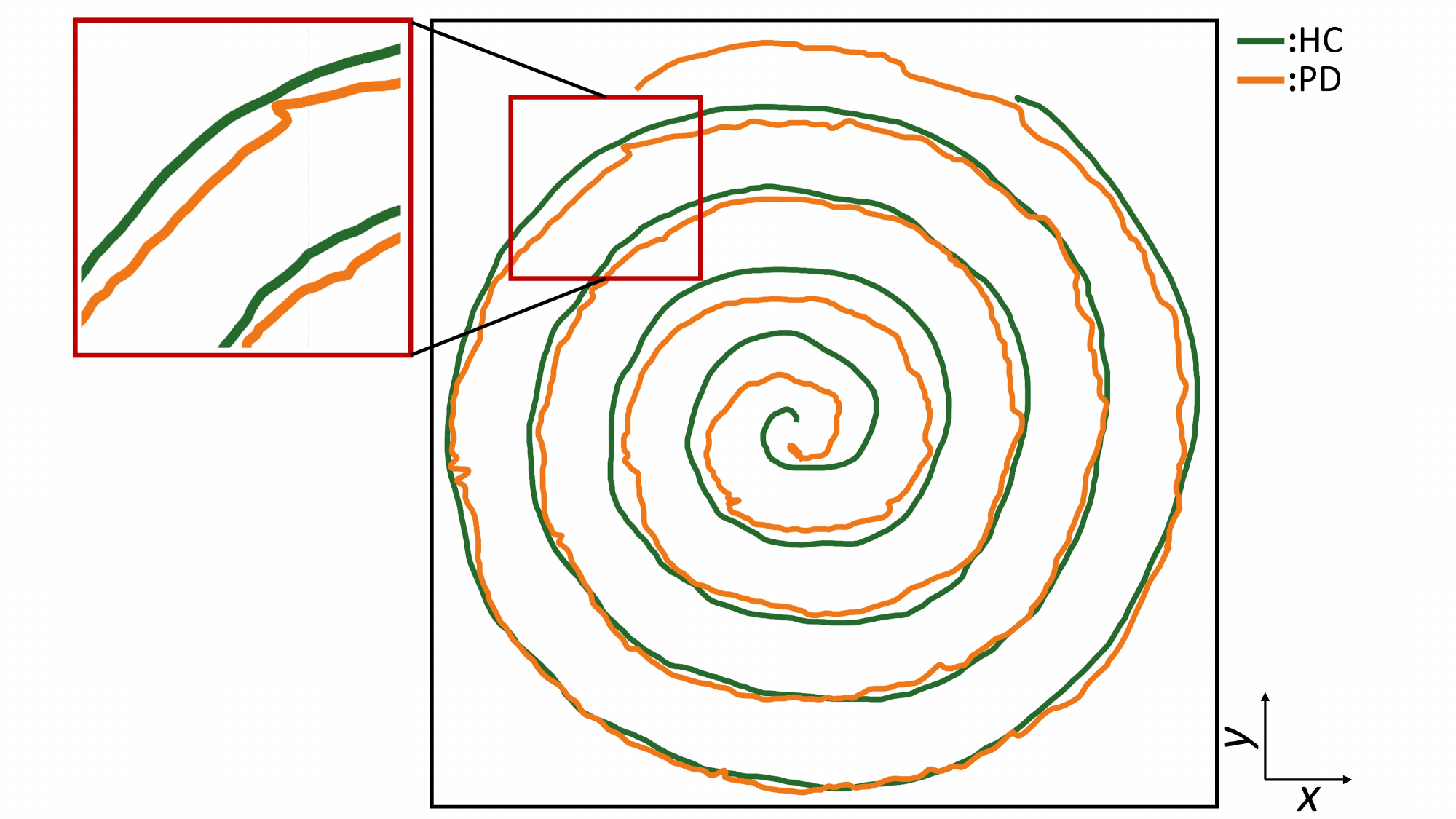}    
        \end{minipage}
    }
    \caption{Comparison of the hand drawings from the Parkinson’s disease (PD) patients and healthy control (HC) subjects, where the relative positions of the hand drawings are reconstructed based on the two-dimensional coordinates of successive hand drawing points over time.}
    \label{Fig.2}
\end{figure*}

More recent, there has been a preliminary exploration of RNN-based models to capture temporal information derived from the temporal dependencies within handwriting signals \cite{ribeiro2019bag}. Furthermore, in \cite{diaz2021sequence}, Diaz et al. propose to use CNNs to extract features from the original feature set and its derived feature set, followed by classification using bidirectional gated recurrent unit (GRU) \cite{cho2014learning}. This approach yield remarkably high diagnostic performance on PaHaW and NewHandPD datasets. While RNN-based networks have not undergone exhaustive exploration in this domain, these studies underscore the potential of $1$D sequence-based dynamic data analysis in early PD diagnosis.

\section{Materials} \label{sec:dataset}
Two datasets have been employed in our study. The first dataset, DraWritePD (acquired by the authors), is used for system fine-tuning to determine the optimal configuration. The second dataset, PaHaW\cite{drotar2014analysis,drotar2016evaluation}, served as an additional test set to evaluate the performance of our method. As illustarated in Fig.\ref{Fig.2}, three distinct sets of handwriting tasks are utilized to validate the robustness of our system. More details of the two datasets are described as follows.

\subsection{DraWritePD}
Data acquisition is carried out with an iPad Pro ($9.7$ inches) equipped with a stylus. $20$ patients meeting the clinical confirmation criteria \cite{goetz2008movement} for PD and $29$ healthy control (HC) subjects gender and age-matched are taken as the control group. Fig.\ref{Fig.2}(a) and Fig.\ref{Fig.2}(b) show the shapes of the \textit{$\Pi$} task and \textit{$\Pi$}\textit{$\Lambda$} task, respectively. During each task, the iPad Pro scans the stylus signal at a fixed rate, and the collected dynamic signal contains approximately $500$-$20000$ data points due to the varying writing speeds of the individuals. For each data point, the dynamic sequence signal captures six sets of independent dynamic variables, including: y-coordinate (mm), x-coordinate (mm), timestamp (sec), azimuth (rad), altitude (rad), pressure (arbitrary unit of force applied on the surface). The data acquisition process is carried out under strict privacy law guidelines. The study is approved by the Research Ethics Committee of the University of Tartu (No.$1275T-9$).  

\subsection{PaHaW}
The PaHaW dataset collects handwriting data from $37$ PD patients and $38$ age and gender-matched HC subjects \cite{drotar2014analysis,drotar2016evaluation}. During the acquisition of PaHaW dataset, each subject is asked to complete handwriting tasks according to the prepared pre-filled template at a comfortable speed. Fig.\ref{Fig.2}(c) shows the shape of the spiral task. Handwriting signals are recorded using a digitizing tablet overlaid with a blank sheet of paper (signals are recorded using an Intuos $4$M pen of frequency $200$ Hz). For each data point, the dynamic signal captures seven sets of independent dynamic variables, including: y-coordinate, x-coordinate, timestamp, button status, altitude, azimuth, pressure. All variables are converted to the same units as in DraWritePD.

\section{Methodology} \label{sec:method}

In this section, we briefly introduce proposed method for automatically diagnosing PD. Fig.\ref{Fig.1} illustrates the schematic workflow of the proposed method, with further details provided in the subsequent subsections.

\subsection{Data Processing} \label{sec:feature engineering}

In this study, we employ several data processing tools, involving mainly normalization, enhancement, and sequence segmentation, to standardise the data. Firstly, we use the Min-Max normalization to ensure that data across different variables shares a similar scale. Note that we only consider five variables, that is, x- and y-coordinates, azimuth, altitude, and pressure for our classification\cite{drotar2014analysis,drotar2016evaluation,impedovo2019velocity}. 

Furthermore, it is worth noting that, in contrast to HC subjects, individuals with PD often exhibit distinctive handwriting patterns during writing tasks. As shown in Fig. \ref{Fig.2}, it is clear in the zoomed-in patches that these PD subjects suffer from more pronounced local tremor than HC subjects. As a result, the extraction of relevant features from handwriting signals has always played a key role for the PD diagnostics \cite{thomas2017handwriting, aouraghe2023literature}. For this purpose, we take a simple but efficient first-order forward difference operation to extract the variation of the handwriting signal in adjacent data points. This strategy helps us to make full use of the local information of handwriting patterns, in particular that the high sampling rate of the acquisition device provides a precise approximation of the actual pen trajectory. To be specific, we apply the forward difference operation only on the x-coordinate and y-coordinate variables individually to accentuate motion patterns pertinent to PD, while keeping the azimuth, altitude, and pressure variables unaltered. Additionally, we also use zero padding to ensure all five individual dynamic variables keeping the uniform size during the process.

\begin{figure}[!t]
    \centering  
    \includegraphics[width=0.482\textwidth]{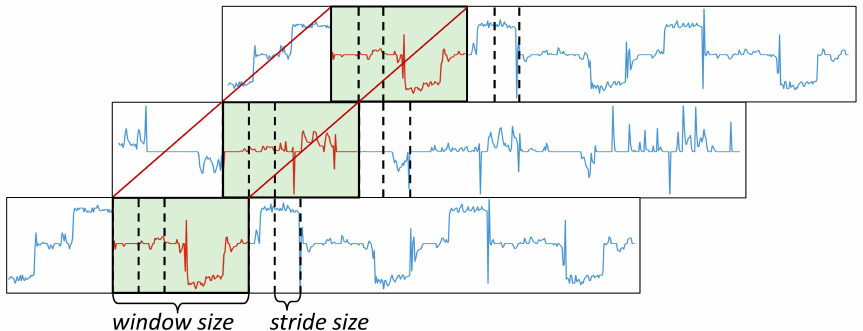}
    \caption{Illustration of temporal data segmentation for given hand-writing sequence, in which the overlap between patches is employed to enrich the data samples. The blue curve represents the independent dynamic variables within the handwriting signal, while the green area depicts the segmented patch data. The \textit{window size} determines the length of the patch data segment, and the \textit{stride size} controls the extent of overlap between adjacent patch data segments.} 
    \label{Fig.3}
\end{figure}

Finally, a data segmentation technique is also applied to preserve local temporal information, where the sequence data is cropped into multiple patches, controlling by two parameters: \textit{window size} ($w$) and \textit{stride size} ($s$). The window size is called as temporal windows \cite{jordao2018human}, which indicates the length of the cropped patches and may vary depending on the characteristics of the data. For simplicity, a fixed window is employed in our experiments. The stride size controls the number of patch samplings in a given sequence, as well as the degree of overlap of patches. Notice also that it is also possible to use a non-uniform stride size during the segmentation step.  The process is also descripted in Fig. \ref{Fig.3}. Moreover, such a segmentation scheme also helps to generate more data samples, which is essential for the subsequent model training.

\subsection{Network Architecture} \label{sec:model}

\begin{figure*}[htbp]
	\centering  
	\includegraphics[width=0.9\textwidth]{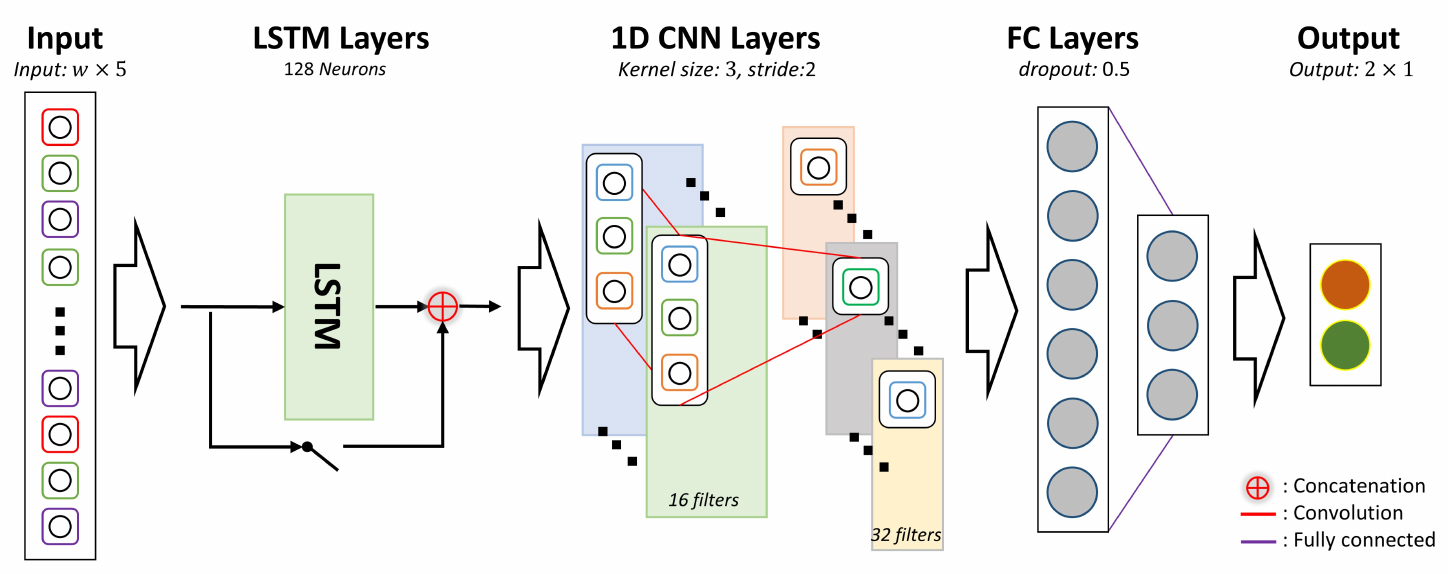}
    	\caption{Network architecture of LSTM-CNN for handwriting signal classification. The LSTM-CNN takes a patch sequence as input, with a length of $w$ and a feature dimension of $5$. The architecture comprises a single LSTM layer with $128$ recurrent units, followed by a concatenation operation that concatenates the input data with the LSTM output features. Subsequently, two one-dimensional convolutional layers, one with $16$ filters and another with $32$ filters, both employing a stride of $2$ and a kernel size of $3$, are applied. Finally, the output layer comprises a single neuron with softmax activation to predict the diagnostic results (PD or HC).} 
	\label{Fig.4}
\end{figure*}

To establish our model architecture, we integrate multiple deep learning structures, including both LSTM \cite{sherstinsky2020fundamentals} and 1D CNN networks. The detailed design of LSTM-CNN is shown in Fig. \ref{Fig.4}, consisting of two essential components: an LSTM block and a 1D CNN block.

\subsubsection{LSTM Block} 
LSTM \cite{sherstinsky2020fundamentals}, a prominent deep learning network, is extensively employed for processing diverse biomedical data, including electroencephalogram (EEG) signals, electrocardiogram (ECG) signals, genetic sequences, and other related domains \cite{xu2022review}. Unlike conventional feed-forward neural networks, LSTM employs internal memory to process incoming inputs and effectively integrates longer temporal signals. Specifically, LSTM comprises a distinctive set of memory units, where current input data and prior states influence the output of the next state. This enables the capture of temporal features from historical information in handwriting signals. In this study, the LSTM block comprises a single LSTM layer, composed of $128$ memory units. Each memory unit is equipped with cells that incorporate input gates, output gates, and forget gates. These gate mechanisms efficiently control the flow of information. With these capabilities, each cell can effectively preserve desired values over extended time intervals. Furthermore, we incorporate an efficient concatenation operation between the LSTM block and the subsequent $1$D CNN block. This operation concatenates the original input with the output of the LSTM block along the feature dimension, before feeding them to the $1$D CNN block, thus enhancing the temporal features.

\subsubsection{CNN Block}  
CNNs have emerged as the powerful tools for a variety of machine learning tasks \cite{alzubaidi2021review}. For example, $2$D CNNs with millions of parameters possess the ability to learn intricate patterns through training on a large-scale database with well-defined labels. However, these approaches may not be feasible in medical scenarios, particularly when dealing with limited availability of medical data. To address this dilemma, $1$D CNNs have recently emerged as a promising approach, demonstrating state-of-the-art performance in biomedical data classification and early diagnosis \cite{kiranyaz20211d}. Moreover, $1$D CNNs and $2$D CNNs share similar network structures, the main distinction lies in that the convolution operation of $1$D CNNs is performed in only one direction. This implies that, given identical conditions (including configuration, network architecture, and hyperparameters), the computational complexity of a $1$D CNN is significantly lower compared to its $2$D counterpart. For example, convolving an input array with dimensions $M$$\times$$M$ by a $K$$\times$$K$ kernel has the $\sim$$\mathcal{O}(M^2K^2)$ computational complexity, whereas in cases of $1$D convolution (with the identical dimensions, $M$ and $K$), the complexity is only $\sim$$\mathcal{O}(MK)$. Therefore, in this study, the CNN block is composed of two $1$D convolutional layers, each of which performs the $1$D convolution with rectified linear unit (ReLU) activation, followed by the $1$D max-pooling layer. Each layer uses multiple filters of identical size to capture information across various temporal scales. More specifically, the first layer uses $16$ filters, with a kernel of size $3$ and stride $2$. The subsequent layer incorporates an increased quantity of $32$ filters with the same kernel of size $3$ and stride $2$. Finally, LSTM-CNN utilizes a fully connected layer to discriminate handwriting signals. The dropout layer temporarily removes nodes from the network with probability $0.5$ during training phase. 

\begin{algorithm}[t]
    \small
    \floatname{algorithm}{Algorithm}
    \renewcommand{\algorithmicrequire}{\textbf{Input:}}
    \renewcommand{\algorithmicensure}{\textbf{Output:}}
    \caption{Majority Voting Algorithm}
    \label{alg:1}
    \begin{algorithmic}[1]
        \REQUIRE {Handwriting sequences $\mathcal{S}\!=\!\{{s_{k}}\}_{k=1}^K$, Label set $\mathcal{L}=\{L_{k}\}_{k=1}^K$, Voting threshold $\alpha$};
        \ENSURE Diagnostic results $\mathcal{T}=\{T_{k}\}_{k=1}^K$;
        \STATE Generate patches set $\mathcal{P}\!=\!\{{p_{i}^{k}}\}_{i=1}^{N_k}$ from $\mathcal{S}\!=\!\{{s_{k}}\}_{k=1}^K$;
        \FOR{$k=1$ to $K$}
            \STATE Set $c=0$,
            \FOR{$i=1$ to $N_k$}
                \STATE LSTM-CNN predicates $\hat{l_i^k}$ for $p_{i}^{k}$,
                \IF {$\hat{l_i^k}=L_k$}
                    \STATE $c = c+1$;
                \ENDIF
            \ENDFOR
            \STATE  Calculate $r_{k}=c/{N_k}$;
            \IF{ $r_{k} \ge \alpha$}
                \STATE $T_{k} = L_{k}$;
            \ELSE
                \STATE $T_{k} \neq L_{k}$;
            \ENDIF
        \ENDFOR  
        \RETURN $\mathcal{T}$;
    \end{algorithmic}  
\end{algorithm}

\subsection{Inference Diagnosis}
Subsequently, we take a majority voting scheme for PD diagnosis. Notice that the LSTM-CNN model is restricted to process input samples with a fixed length $w$, we start to segment the acquired handwriting signals $\mathcal{S}=\{{s_{k}},k=1,2,\dots,K\}$ into a set of fixed-length patches $\mathcal{P}=\{{p_{i}^{k}},i=1,2,\dots,N_k\}$ based on the aforementioned data segmentation technique. Once the patches data are obtained, the LSTM-CNN model predicates the corresponding classification results for these patches. For each sequence data $s_{k}$, a majority voting scheme is adopted to determine the ultimate inference result. For example, given a threshold $\alpha \in (0,1)$ and true label $L_k$, the percentage $r_k$ represents the correctly classified patches. If $r_k \ge \alpha$, the sequence $s_{k}$ is correctly predicated with corresponding predicted label $T_k$ as $L_k$, otherwise it is not. In general, unless otherwise stated in the subsequent experiments, we set the threshold $\alpha = 0.5$ to indicate that the voting scheme should be consistent with the majority reliable predictions. More details of choosing $\alpha$ is also discussed in Section \ref{sec:ablation}. The majority voting scheme are presented in Algorithm\ref{alg:1}.

\section{Experiments}\label{sec:results}

In this section, we present quantitative experiments to evaluate the performance of the proposed LSTM-CNN model. The five-fold cross-validation technique is deduced in the context of the handwriting tasks. For empirical analysis, we adopt four metrics, including accuracy, recall, $\text{F}_{1}$ score, and Matthews correlation coefficient (MCC) \cite{grandini2020metrics}, which have been widely-used in contemptuous classification tasks. All experiments are implemented on a desktop PC equipped with an Intel i$7$-$11700$K $3.60$ GHZ, $32$GB RAM and an $8$ GB Nvidia RTX$3070$Ti GPU.

\begin{table*}[!t]
    \small
    \renewcommand{\arraystretch}{1.1}
    \caption{Quantitative comparison with various classification models on the DraWritePD dataset.} 
    \label{Tab.1} 
    \setlength{\tabcolsep}{1.5mm}{
    \begin{tabular}{l | c | c | c | c | c | c | c | c}
        \toprule
        \multirow{2}*{\diagbox[width=7em]{Metric}{Method}} & \multicolumn{4}{c|}{Model-based methods} &  \multicolumn{4}{c}{Deep learning-based methods} \\
       \cline{2-9}
        & KNN & SVM & Adaboost & RF & CNN & RNN-CNN & GRU-CNN & LSTM-CNN(Ours) \\
        \hline \hline
        Recall ($\%$) & $81.8$ / $72.7$ & $87.4$ / $79.2$ & $93.0$ / $81.8$ & $93.0$ / $84.3$ & $80.0$ / $83.6$ & $83.6$ / $85.5$ & $89.1$ / $87.3$ & $\bm{94.5}$ / $\bm{89.1}$ \\
      Accuracy ($\%$) & $92.0$ / $84.6$ & $93.6$ / $87.7$ & $93.6$ / $88.5$ & $93.6$ / $86.2$ & $91.5$ / $92.8$ & $92.8$ / $93.1$ & $94.6$ / $94.4$ & $\bm{96.2}$ / $\bm{95.2}$ \\
      $\text{F}_{1}$ score ($\%$) & $90.0$ / $80.0$ & $93.3$ / $86.9$ & $92.6$ / $85.7$ & $92.6$ / $82.3$ & $88.7$ / $91.0$ & $91.0$ / $90.9$ & $93.1$ / $93.1$ & $\bm{95.4}$ / $\bm{94.2}$ \\
       MCC & $0.85$ / $0.69$ & $0.89$ / $0.89$ & $0.87$ / $0.76$ & $0.87$ / $0.72$ & $0.84$ / $0.86$ & $0.86$ / $0.87$ & $0.89$ / $0.89$ & $\bm{0.92}$ / $\bm{0.91}$ \\
       \hline 
        Params (K) & -- & -- & -- & -- & $42.69$ & $83.71$ & $83.83$ & $83.89$ \\
        FLOPs (K) & -- & -- & -- & -- & $202.37$ & $446.21$ & $542.21$ & $590.21$ \\  
        \toprule
    \end{tabular} }
\end{table*}



\subsection{Comparison with existing methods}

We compare in Table \ref{Tab.1} our model with methods widely adopted \cite{thomas2017handwriting, aouraghe2023literature} for PD diagnosis on the DraWritePD dataset. All diagnostic methods are classified into model-based and deep learning-based approaches, according to the differences in model architecture. For model-based methods, we implement them in Python using the Scikit-learn library \cite{pedregosa2011scikit}. Additionally, the grid search algorithm is employed to optimize hyperparameters. Specifically, (i) k-Nearest Neighbors (KNN): the possible number of neighbors is $K=\left[3,5,7,9,11\right]$. (ii) SVM: the radial basis function kernel is employed, the optimization range of kernel gamma $\gamma$ and penalty parameter $C$ are $\gamma=\left[2^{-2},2^{-1},2^0,2^1,2^2\right]$ and $C=\left[2^{-3},2^{-2},2^{-1},2^0,2^1\right]$, respectively. (iii) Adaboost and RandomForest(RF): the possible number of decision trees $N$ and the optimization range of maximum depth $D$ are $N=\left[5,10,15,20,50\right]$ and $D=\left[2,4,8,16,32\right]$, respectively. On the other hand, for deep learning-based methods, we implement them using the Pytorch framework \cite{paszke2019pytorch}. Specifically, (i) CNN: the architecture is a standard and compact neural network, comprising solely of two $1$D convolutional layers and fully connected layers. (ii) RNN-CNN and GRU-CNN: their architectures contain a layer of memory units stacked on top of the aforementioned CNN, and the only distinction from LSTM-CNN lies in the variation of memory units. 

\begin{figure}[t]
	\centering  
	\includegraphics[width=0.43\textwidth ]{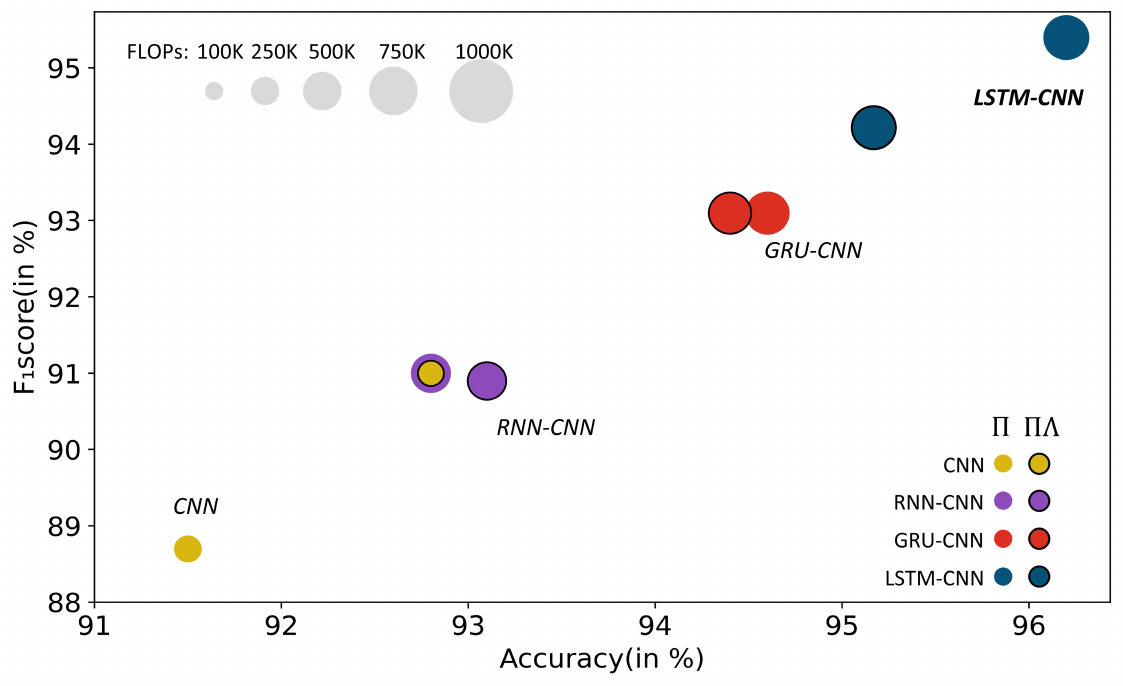}
        \caption{The trade-off between performance and efficiency: our method v.s. comparison methods.} 
	\label{Fig.5}
\end{figure}

For comparison, we also report the results of both model-based and other deep learning-based approaches. It is worth noting that pairwise results in the table are composed sequentially of the results for the \textit{$\Pi$} task and the \textit{$\Pi$}\textit{$\Lambda$} task. From this table, we observe following: (i) Our model outperforms the existing model-based methods significantly. This demonstrates the effectiveness of our approach to exploiting neural network architectures for PD diagnosis. (ii) Our model gives superior results among deep learning-based methods. Also, the performance enhancements, especially in the \textit{$\Pi$} task, are particularly significant when compared to other hybrid architectures incorporating memory units. This confirms the effectiveness of such a hybrid architecture, where memory units, especially LSTM, contribute to more discriminative temporal features. (iii) While other compact deep learning-based architectures are slightly smaller in size than ours, they are outperformed by our model by a significant margin in terms of all performance metrics. Note that our model is lightweight enough to compete with other established ones such as Alexnet \cite{krizhevsky2012imagenet}, Resnet \cite{he2016deep}, Transfomer \cite{vaswani2017attention}.

\begin{table}[!t]
    \small
    \renewcommand{\arraystretch}{1.2}
    \caption{Time consumption of LSTM-CNN at each inference diagnosis stage (in seconds).}
    \label{Tab.2}
    \setlength{\tabcolsep}{1.5mm}{
        \begin{tabular}{c | c | c c c | c}
            \toprule
             Category & Size & Loading & Processing & Model & Total \\
             \hline \hline
             HC & $667$ & 0.08491 & 0.00197 & 0.01097 & 0.10645 \\
             PD & $3128$ & 0.09081 & 0.00299 & 0.01494 & 0.14963 \\
             HC & $7122$ & 0.09108 & 0.00897 & 0.01795 & 0.16689 \\
             PD & $9471$ & 0.09984 & 0.01097 & 0.02294 & 0.18761 \\
             PD & $12787$ & 0.08614 & 0.01696 & 0.02690 & 0.19087 \\
             PD & $18618$ & 0.09197 & 0.01995 & 0.03291 & 0.21960 \\
            \toprule
        \end{tabular} }
\end{table}

\begin{figure*}[!b]
    \centering
    \subfigure[input features]{
        \includegraphics[width=0.22\textwidth]{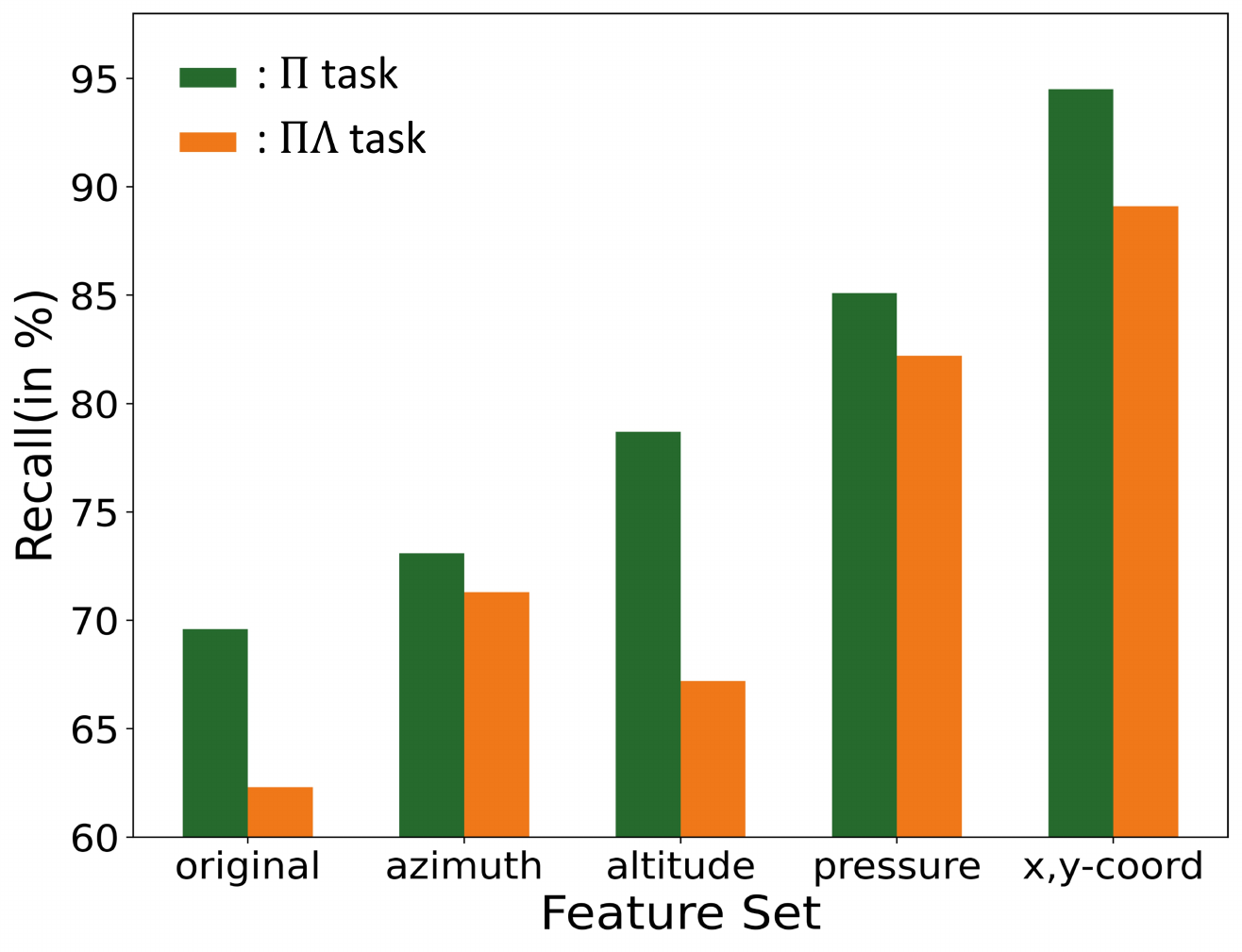}
    }
    \subfigure[input features]{
        \includegraphics[width=0.22\textwidth]{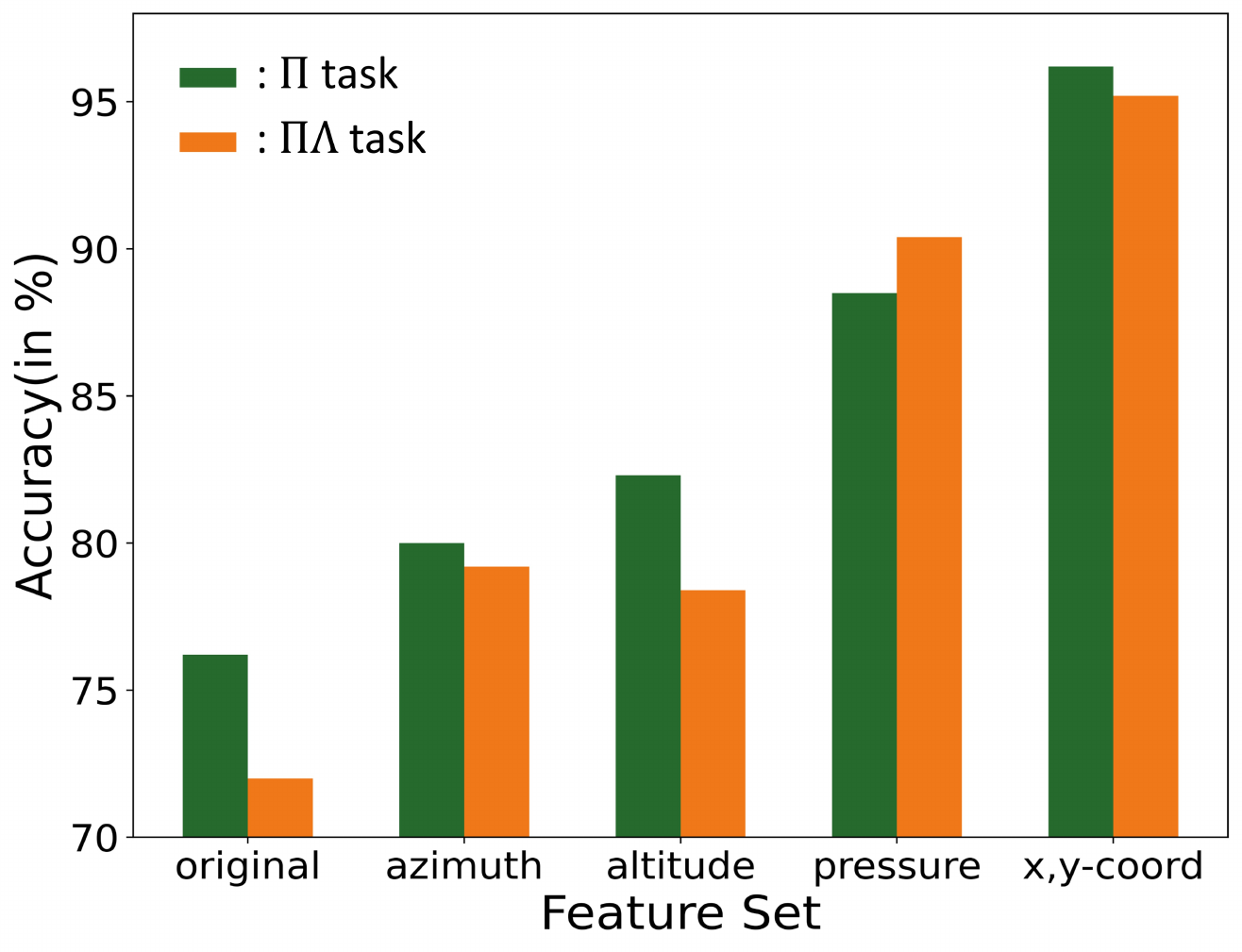}
    }
    \subfigure[input features]{
        \includegraphics[width=0.22\textwidth]{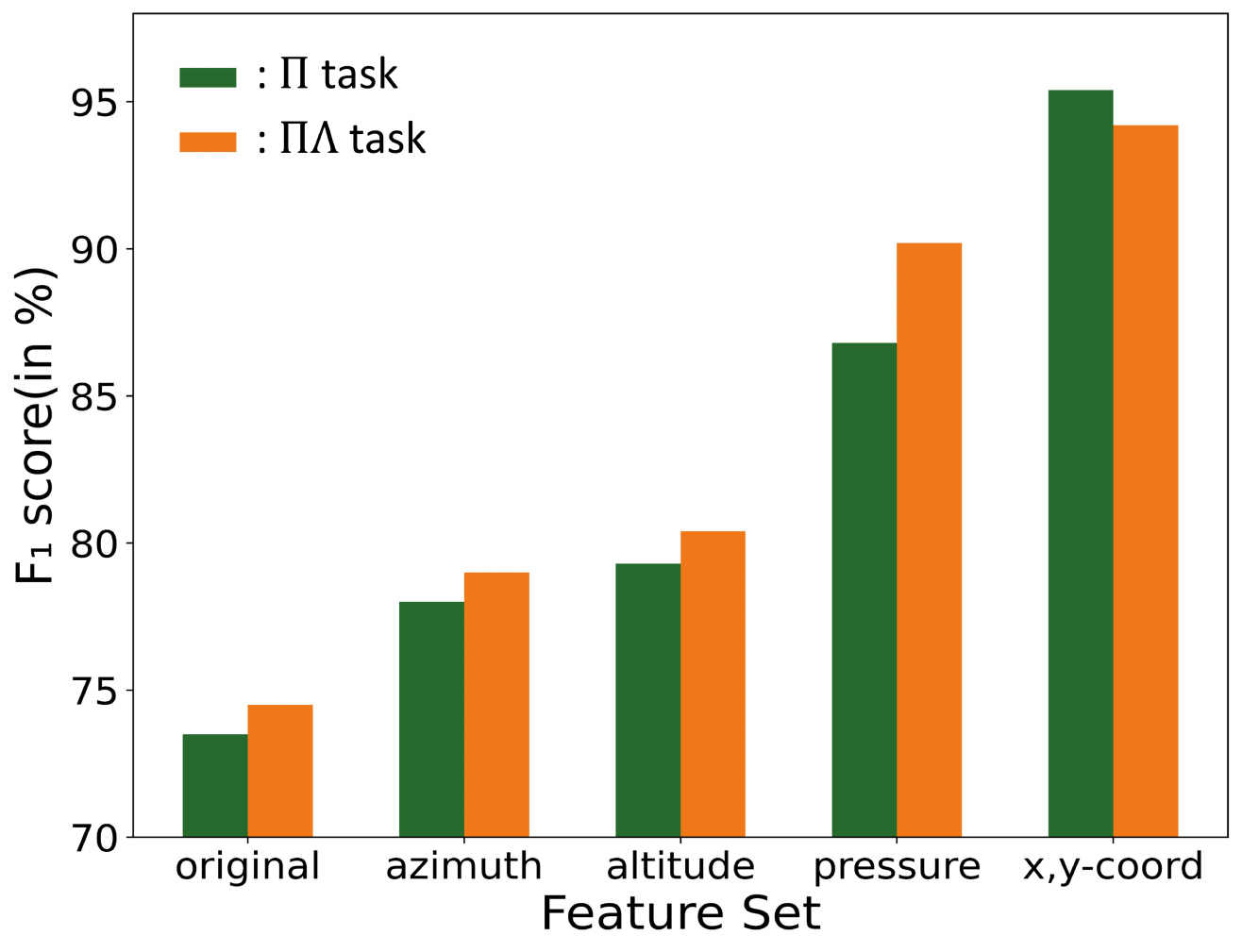}
    }
    \subfigure[input features]{
        \includegraphics[width=0.22\textwidth]{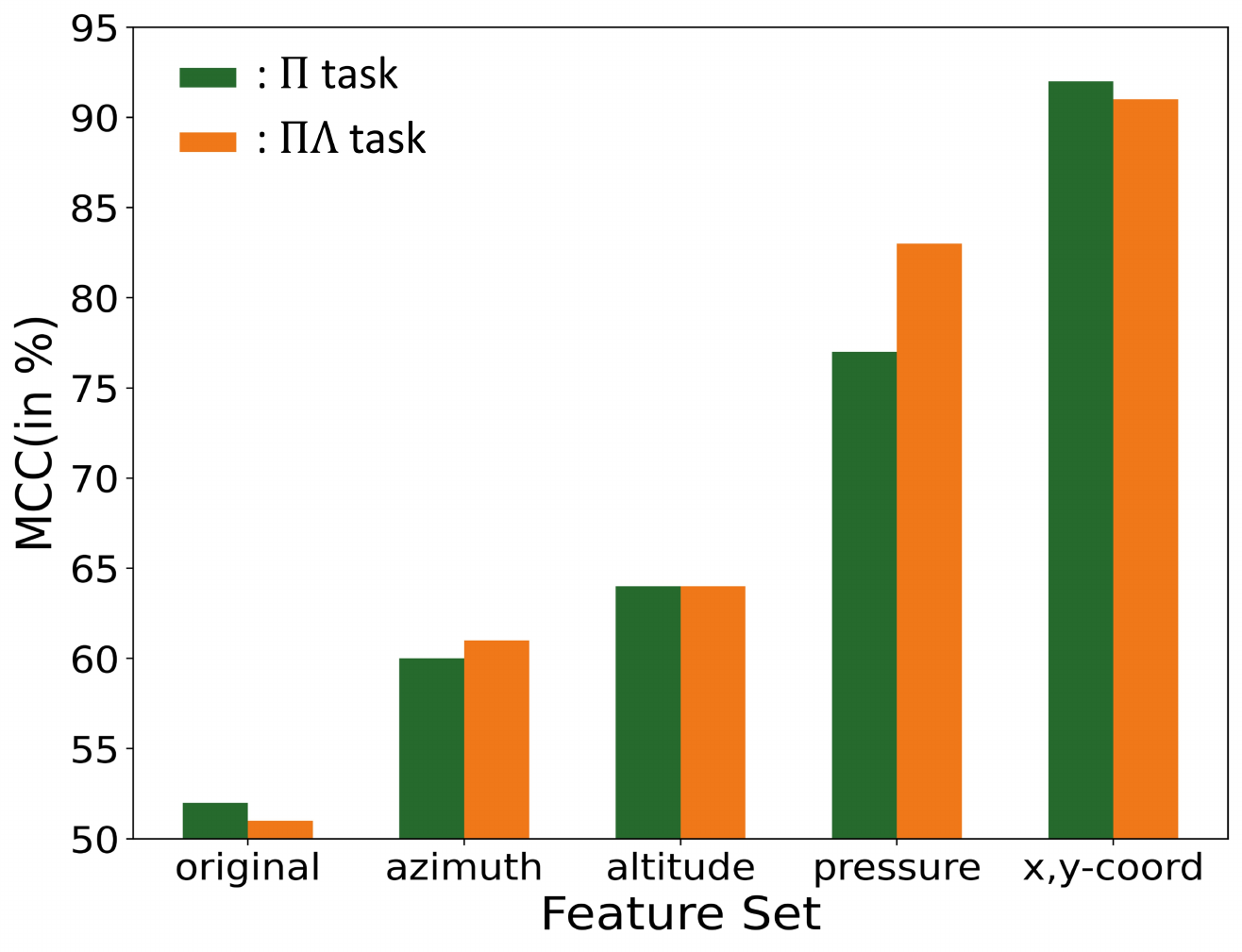}
    }
    \subfigure[temporal window]{
        \includegraphics[width=0.22\textwidth]{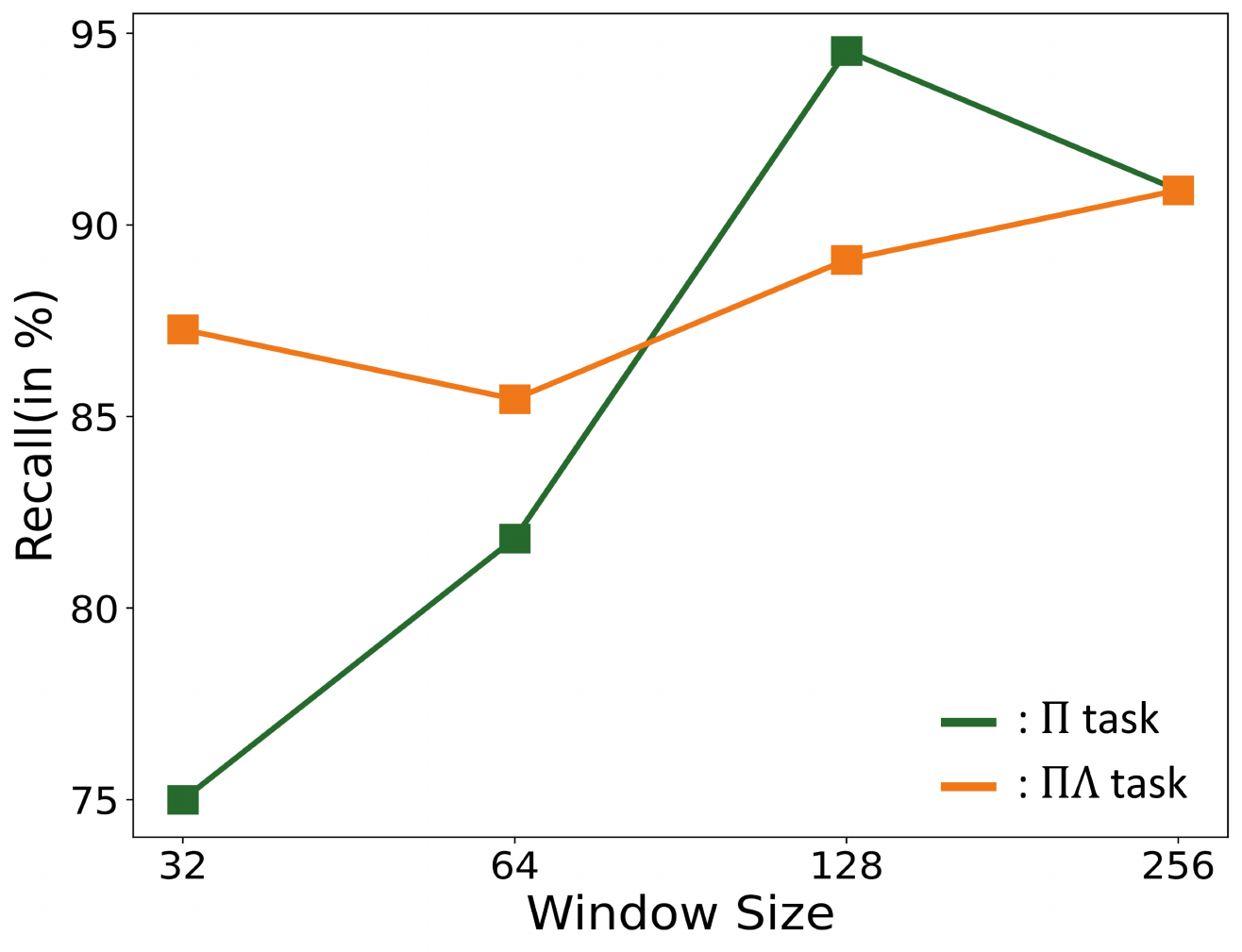}
    }
    \subfigure[temporal window]{
        \includegraphics[width=0.22\textwidth]{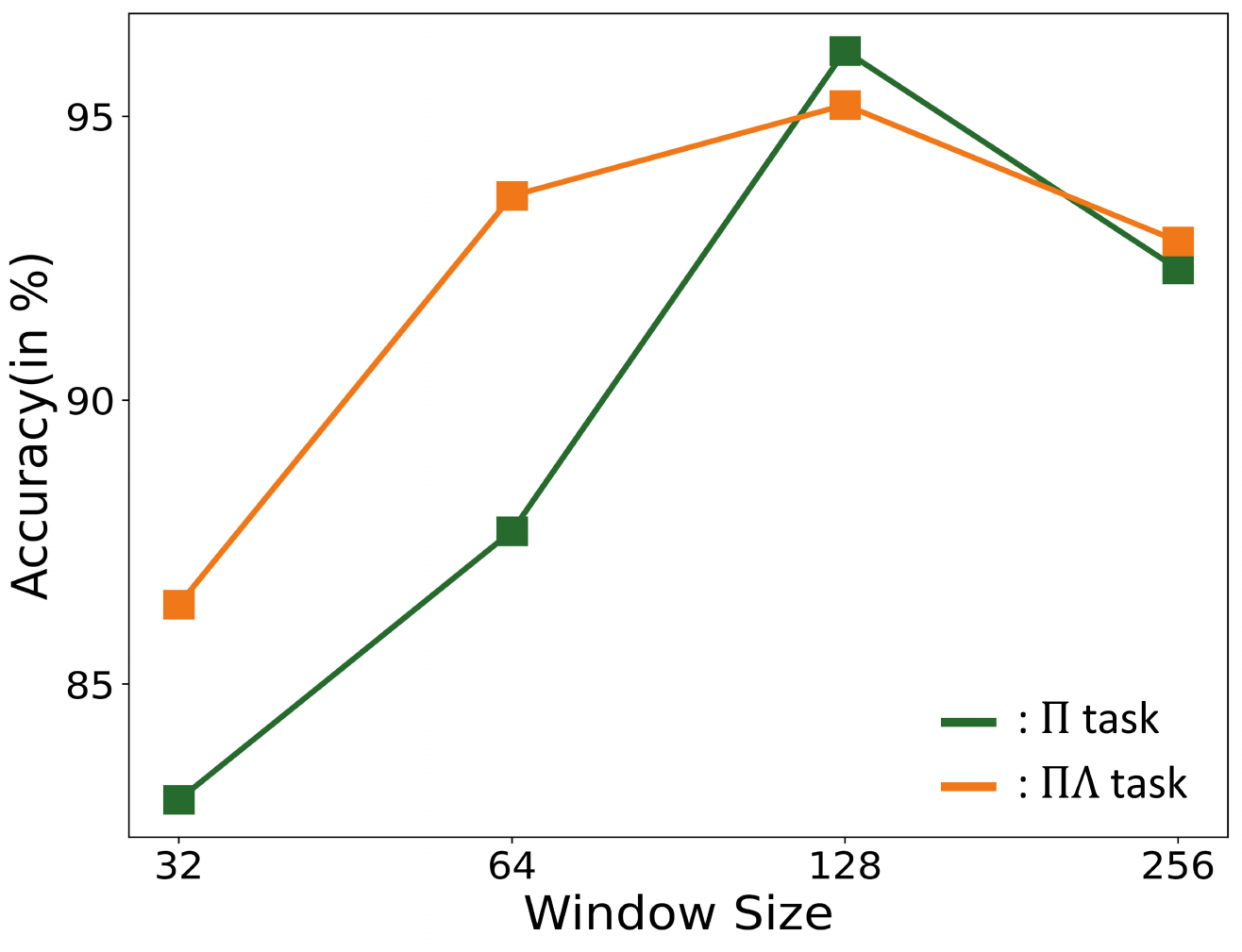}
    }
    \subfigure[temporal window]{
        \includegraphics[width=0.22\textwidth]{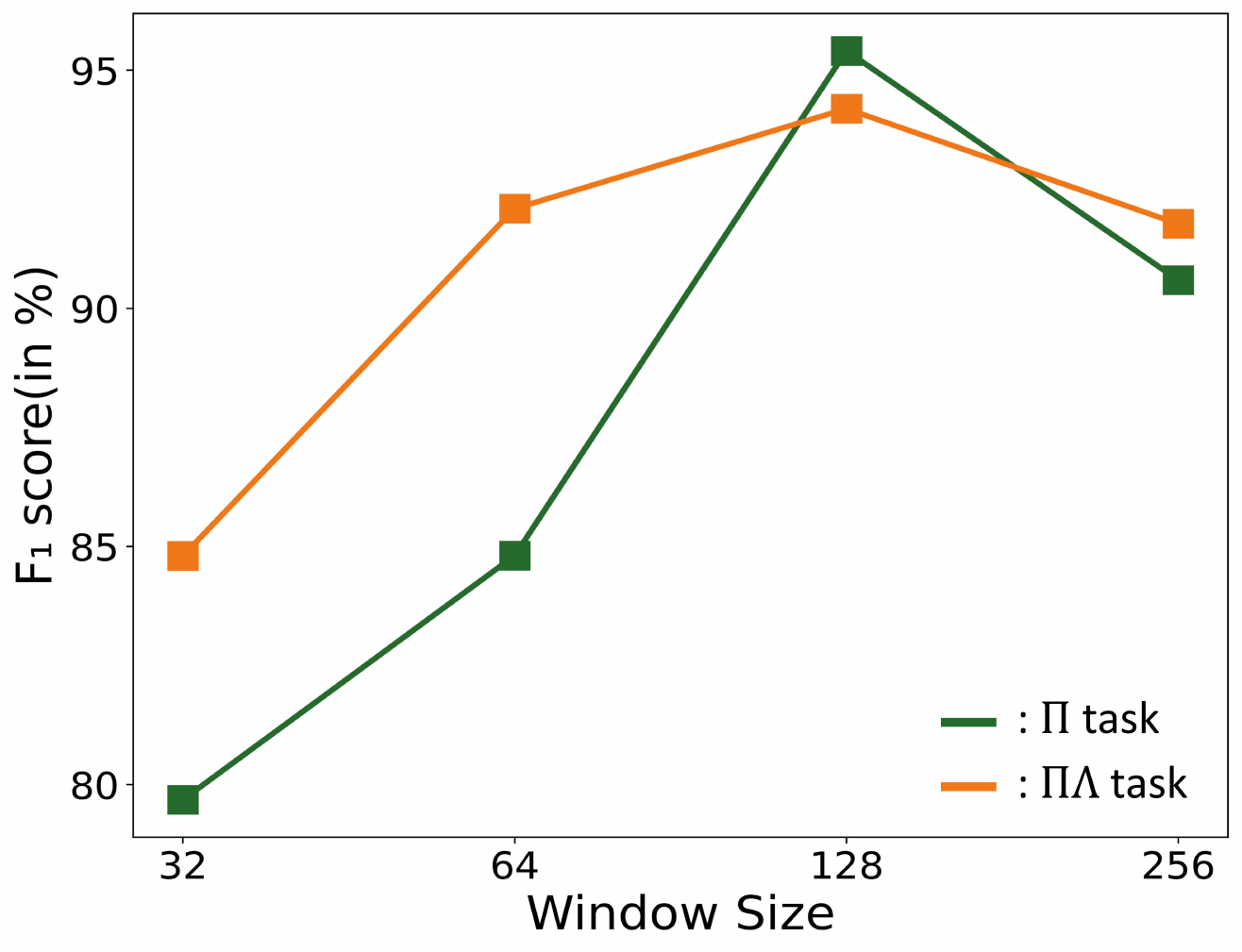}
    }
    \subfigure[temporal window]{
        \includegraphics[width=0.22\textwidth]{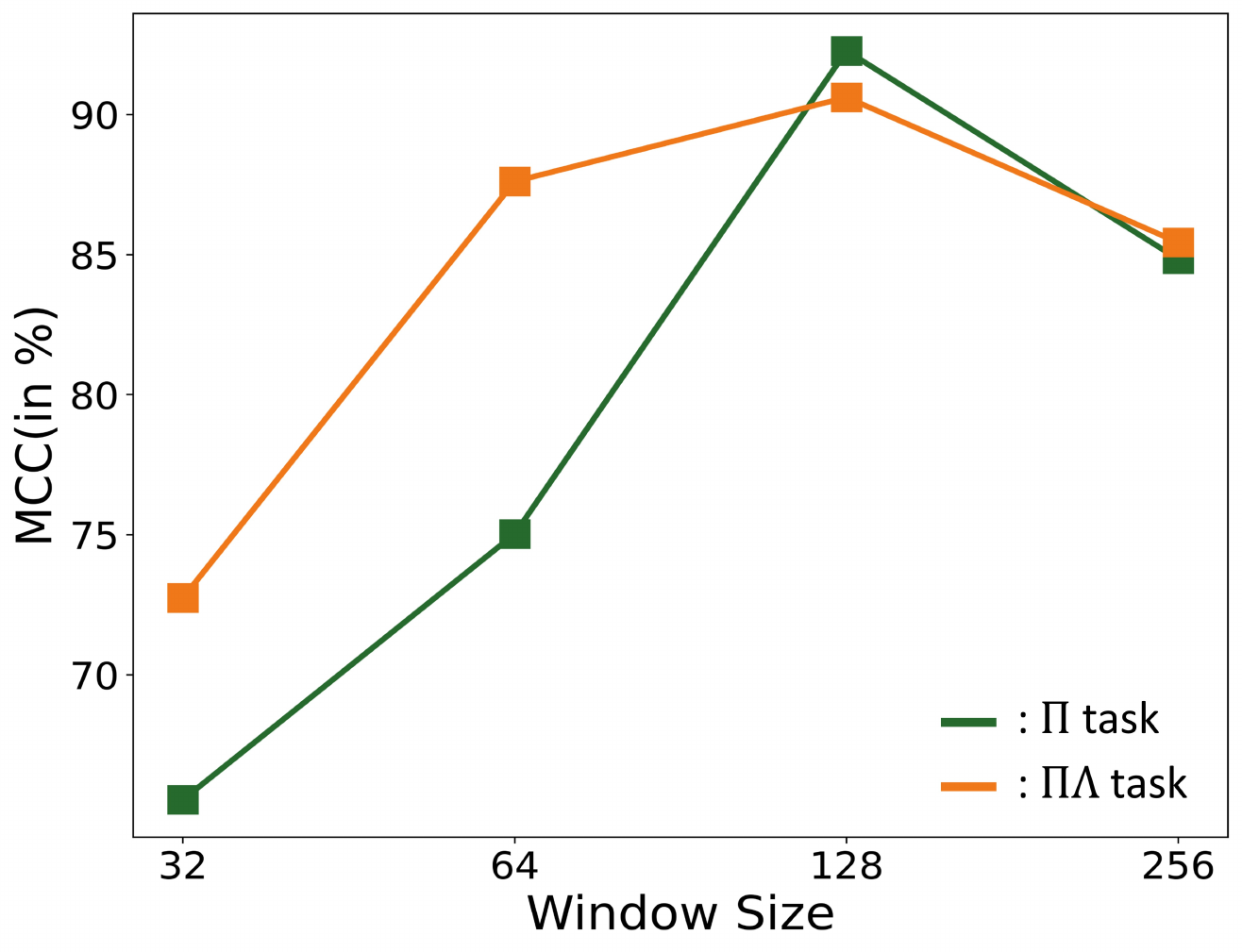}
    }
    \subfigure[threshold]{
        \includegraphics[width=0.22\textwidth]{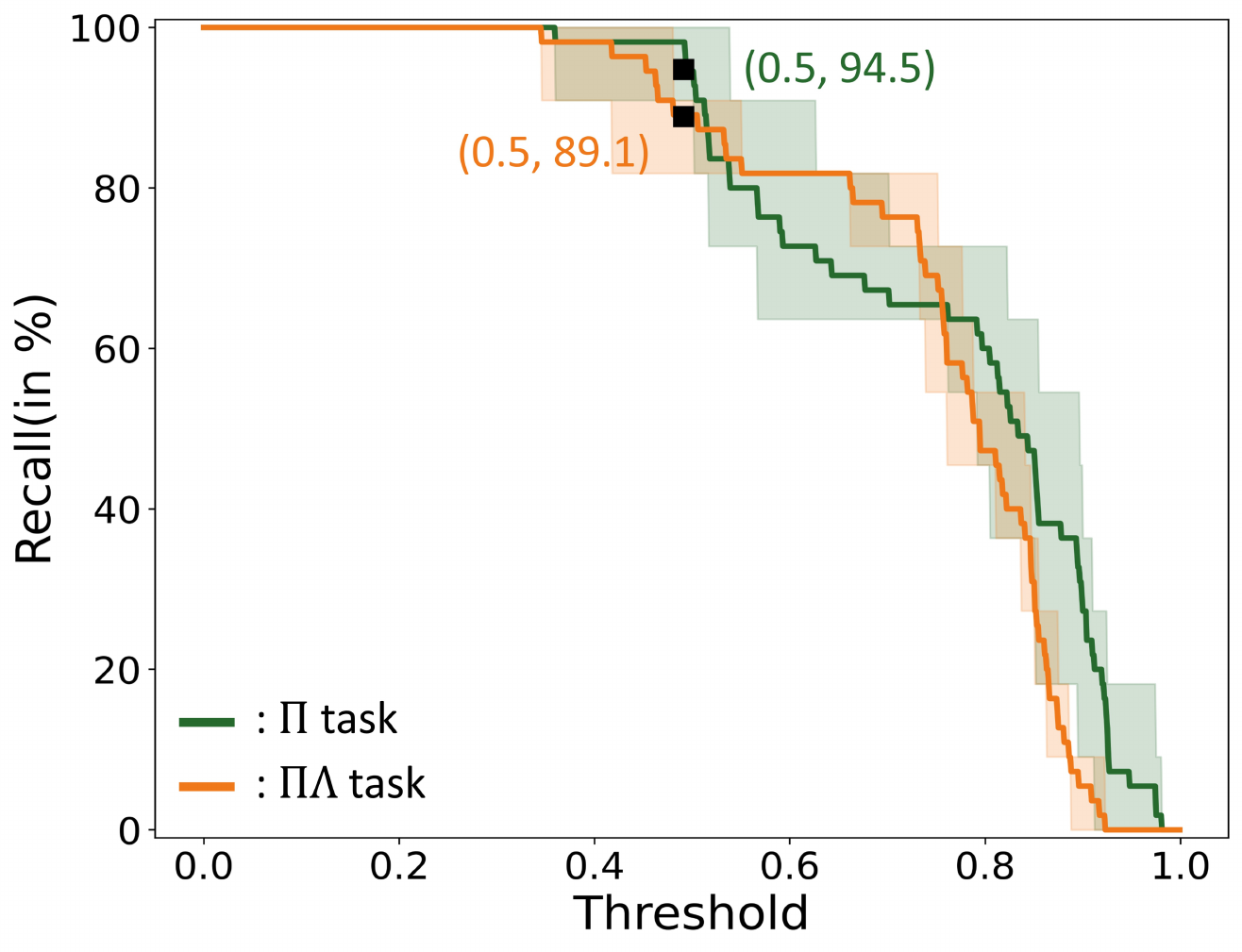}
    }
    \subfigure[threshold]{
        \includegraphics[width=0.22\textwidth]{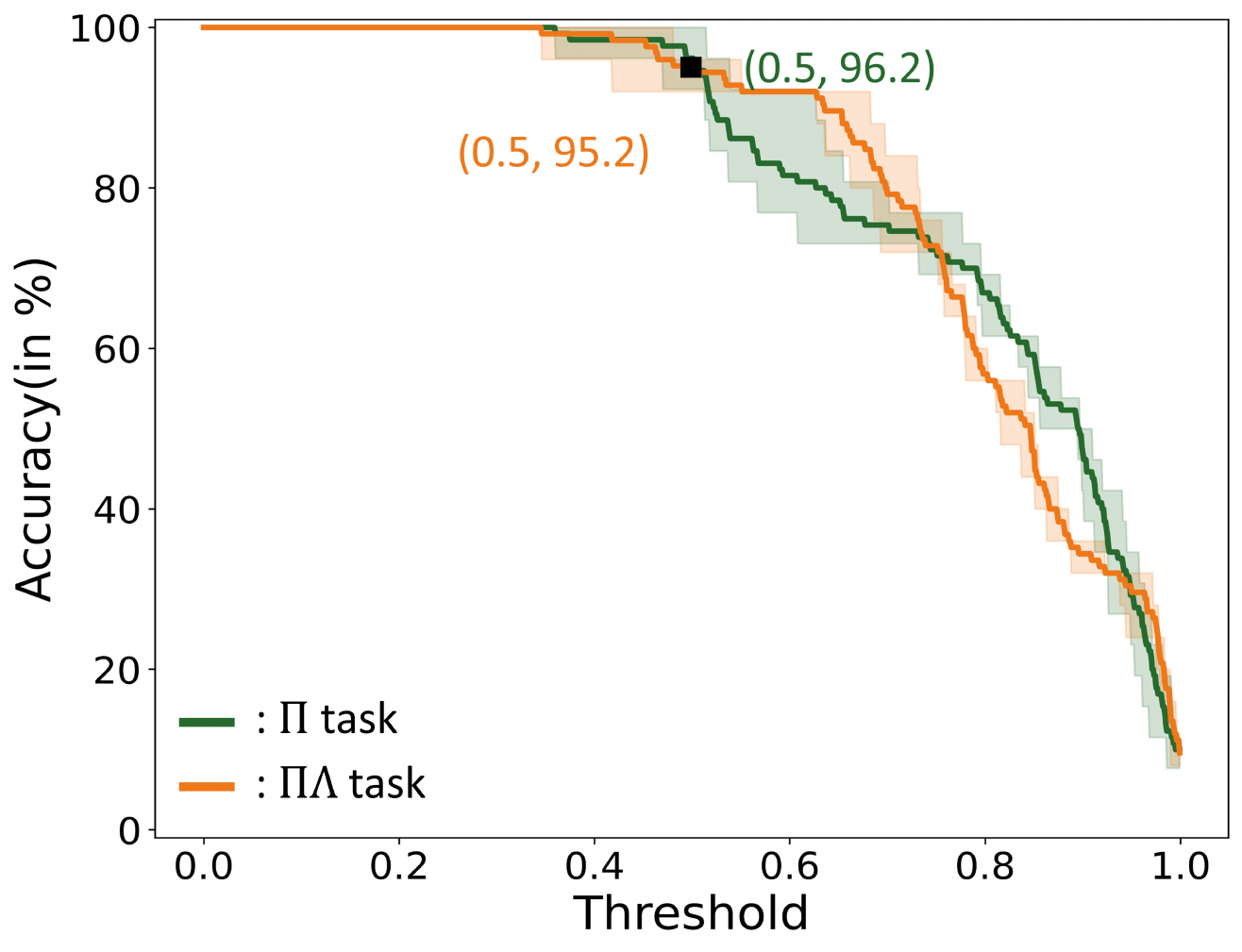}
    }
    \subfigure[threshold]{
        \includegraphics[width=0.22\textwidth]{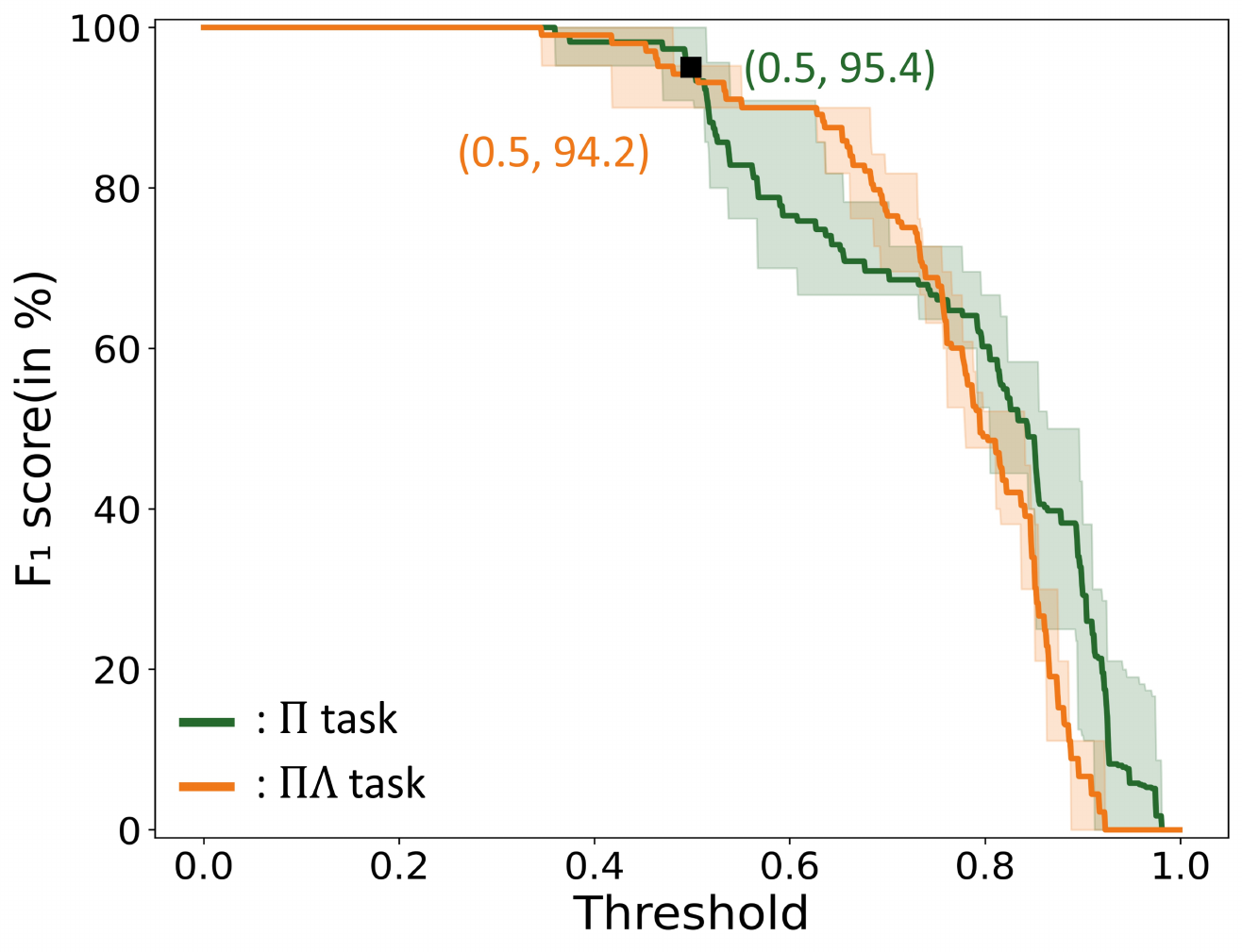}
    }
    \subfigure[threshold]{
        \includegraphics[width=0.22\textwidth]{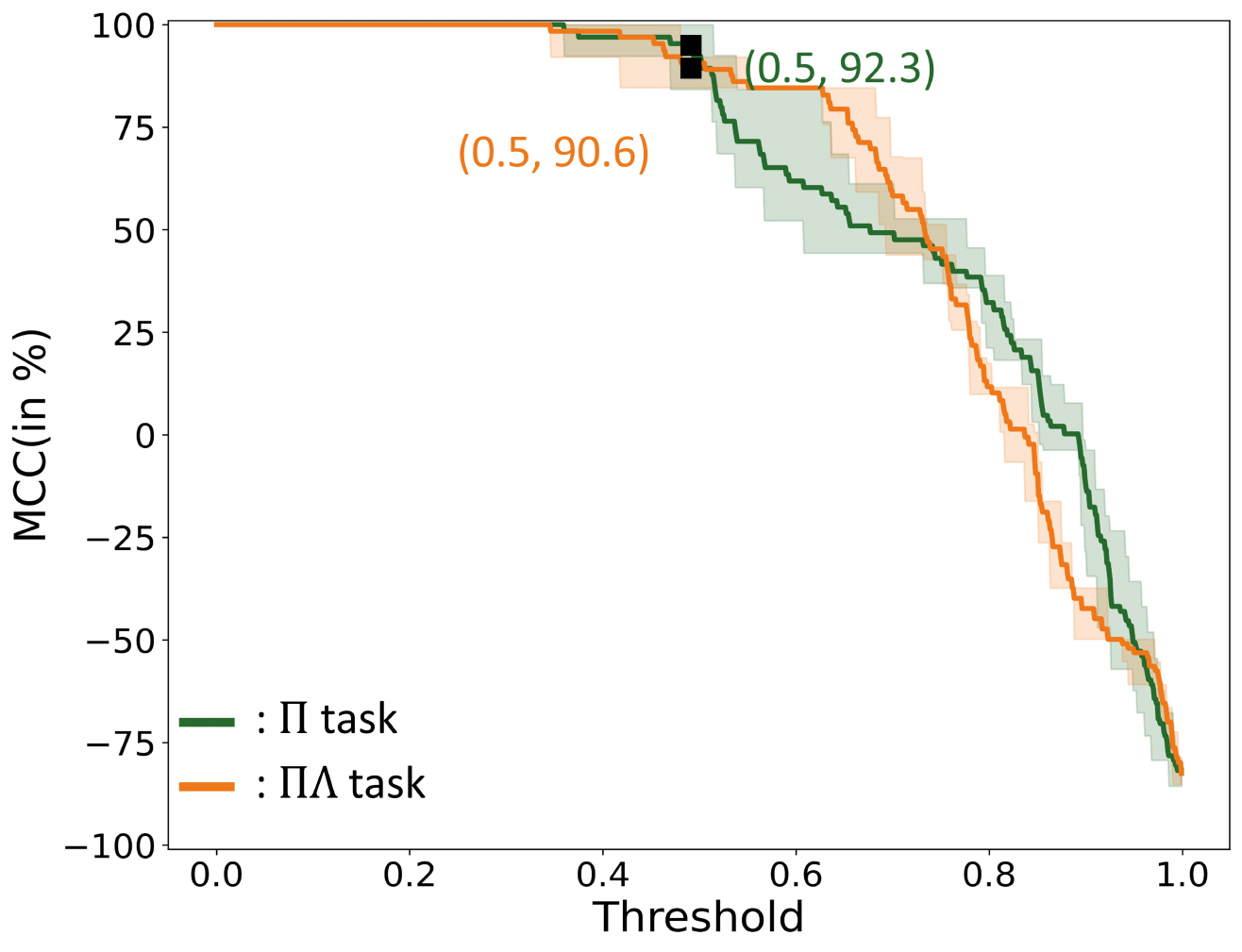}
    }
    \caption{Quantitative comparison of various configurations: (a)-(d) evaluating the influence of input feature sets, (e)-(h) comparing model performance under different window lengths, and (i)-(l) illustrating model performance as a function of diagnostic threshold.}
    \label{Fig.6}
\end{figure*} 

\subsection{Complexity Analysis}

We further evaluate the efficiency of the proposed LSTM-CNN from the perspectives of both computational complexity and inference time. As presented in Table \ref{Tab.1}, the LSTM-CNN exhibits an impressively low parameter count of only $0.084$ million, accompanied by a mere total of $0.59$ million floating point operations (FLOPs). This theoretically indicates that the model possesses exceptional lightweight characteristics and boasts an exceptionally low computational complexity. Furthermore, we also integrate diagnostic performance metrics with computational complexity to conduct a comprehensive evaluation. As illustrated in Fig.\ref{Fig.5}, it is evident that the LSTM-CNN demonstrates significantly superior diagnostic performance while maintaining an acceptable level of computational complexity. 

On the other hand, we also conduct on-site monitoring of the actual runtime duration for various stages involved in the inference diagnosis. Table \ref{Tab.2} primarily presents handwriting sequence scale, model prediction duration, and the overall duration. We can conclude that the overall diagnosis duration does not exceed $0.3$ seconds, while the model prediction is almost real-time, taking only $0.03$ seconds. This observation further substantiates the lightweight and efficient nature of the proposed LSTM-CNN. Note that the algorithm framework is not specifically optimized in this study, and GPUs are not used in the inference diagnostic tests.



\begin{table}[!b]
    \small
    \renewcommand{\arraystretch}{1}
    \caption{The influence of concatenation operation on model performance.}
    \label{Tab.3}
    \setlength{\tabcolsep}{5.6mm}{
    \begin{tabular}{l | c c  }
        \toprule  
         Concatenation & \ding{55} & \ding{51} \\
         \hline \hline
         Recall ($\%$) & $90.9$ / $\bm{89.1}$ & $\bm{94.5}$ / $\bm{89.1}$ \\
         Accuracy ($\%$) & $93.8$ / $\bm{95.2}$ & $\bm{96.2}$ / $\bm{95.2}$ \\
         $\text{F}_{1}$ score ($\%$) & $92.5$ / $94.1$ & $\bm{95.4}$ / $\bm{94.2}$ \\
         MCC & $0.88$ / $\bm{0.91}$ & $\bm{0.92}$ / $\bm{0.91}$ \\
        \toprule 
    \end{tabular}}
\end{table}

\begin{table*}[!t]
    \small
    \renewcommand{\arraystretch}{1}
    \caption{Quantitative comparison with state-of-the-art works on the PaHaW dataset.} 
    \label{Tab.4}
    \setlength{\tabcolsep}{3.2mm}{
        \begin{tabular}{l | c | c | c | c}
            \toprule
            Method & Features & Models & Accuracy ($\%$) & Year \\
            \hline \hline
            Drot$\acute{a}$r et al. \cite{drotar2016evaluation} &  kinematic, spatio-temporal and pressure features & RF, SVM & $62.8$ & $2016$ \\
            Angelillo et al.\cite{angelillo2019performance} & velocity-based features  & SVM & $53.8$ & $2019$ \\ 
            Diaz et al. \cite{diaz2019dynamically} & static images with dynamically enhanced & $2$D CNN+SVM & $75.0$ & $2019$ \\
            Valla et al. \cite{valla2022tremor} & derivative-based, angle-type, and integral-like features & KNN, SVM & $84.9$ & $2022$ \\
            \hline
            Ours & dynamic handwriting patterns & LSTM-CNN & $\bm{90.7}$ & -- \\
            \toprule
        \end{tabular} }
\end{table*}

\subsection{Ablation Studies}\label{sec:ablation}



\paragraph{Design of Input Features.} We conduct comparative experiments to evaluate the impact of the input feature set on the model performance. Specifically, we construct four distinct input feature sets by applying the forward difference operation to each individual variable while keeping the remaining variables constant. It is important to emphasize that the (x-coordinate and y-coordinate) variables collectively determine the coordinates of the data point, hence, they are considered as a single variable. In addition, we include the original variables as one additional input feature set, serving as the baseline. As illustrated in Fig. \ref{Fig.6} (a)-(d), the optimal performance is achieved when geometric variables (i.e. x-coordinate and y-coordinate) undergo the difference operation. The primary reason for this phenomenon, we speculate, lies in that individuals with PD typically exhibit more prominent localized tremors and sustained acceleration peaks during writing assessments in comparison to HC subjects \cite{jerkovic2019analysis}.

\paragraph{Design of Temporal Window.} Given that motor symptoms in patients with PD primarily manifest through local information obtained from the handwriting signal, the careful selection of an appropriate window length ($w$) becomes imperative. In this study, We select four different window lengths for comparison, taking into account the distribution of signal lengths in the datasets. Upon observing Fig.\ref{Fig.6} (e)-(h), we can see that longer window lengths are more likely to yield superior diagnostic performance. In addition, there is a clear trend of increasing and then decreasing curves for most metrics, with the optimal performance achieved at $w=128$.




\paragraph{Design of Concatenation.} In Table \ref{Tab.3}, we compare the impact of the ``concatenation'' in the model architecture. It is worth noting that pairwise results in the table are composed sequentially of the results for the \textit{$\Pi$} task and the \textit{$\Pi$}\textit{$\Lambda$} task. From this table, we can observe that the incorporation of the ``concatenation'' operation, not only maintains stable model performance in the \textit{$\Pi$}\textit{$\Lambda$} task, but also yields an additional improvement of $2\%$ to $4\%$ in model performance for the \textit{$\Pi$} task.

\paragraph{Design of Threshold.} As illustrated in Fig.\ref{Fig.6} (i)-(l), we conduct additional supplementary experiments on the threshold $\alpha$ in the majority voting scheme during the diagnostic inference process, and we annotate the specific performance of each metric when the threshold is set to $0.5$.

\subsection{Robustness Validation} 
In Table \ref{Tab.4}, we use the novel publicly available dataset PaHaW \cite{drotar2014analysis,drotar2016evaluation} to validate the robustness of proposed LSTM-CNN and to conduct a fair comparison with state-of-the-art methods. This dataset is not used in the configuration of our system, therefore, we adopt the optimal configuration described in Section \ref{sec:ablation}. In addition, we analyze previous literature that uses this specific dataset to contextualize our results. It should be noted that various classifiers, such as SVM, RF, and CNN, are used in these studies, resulting in different classification performances. To ensure a fair comparison, only the best results from each study are presented here. As evident from this table, our method exhibits superior performance when compared to previously proposed methods that rely on traditional handcrafted features or $2$D image recognition. This further confirms the effectiveness of the proposed method as a candidate solution for practical use in clinical settings. In addition, it should note that, in our work, we primarily use this dataset to validate the robustness of our method.


\section{Conclusions}\label{sec:conlusions}

The accumulation of evidence \cite{thomas2017handwriting, aouraghe2023literature} derived from dynamic handwriting analysis provides robust support for the hypothesis that distinctive motor patterns of individual can be captured through the analysis of dynamic signals associated with handwriting. In particular, handwriting impairment in patients with PD has been clinically demonstrated, this analysis is expected to help in the diagnosis of PD \cite{ aouraghe2022literature}. To this end, in this paper, we propose a efficient hybrid model that integrates LSTM and $1$D CNN to identify unique patterns in dynamic handwriting sequences. Systematic experimental results substantiate that the proposed model offers efficient diagnostic performance, minimal computational requirements, and strong robustness compared to current state-of-the-art methods. Furthermore, it is worth noting that our study integrates two distinct datasets encompassing various handwriting tasks, with the aim of maximizing the validation of the robustness of our proposed method. A significant limitation of this study is the small size of the dataset we used, which may somewhat affect the generalizability of the obtained results. Despite these limitations, the reported performance values demonstrate significant potential, and it is anticipated that the findings of this study will pave the way for an operational system in a clinical setting.


\section*{Acknowledgements}
This study was supported by the Grant PRG$957$ of the Estonian Research Council. This work in the project ``ICT programme'' was supported by the European Union through the European Social Fund. It was also partially supported by the FWO Odysseus 1 grant G.0H94.18N: Analysis and Partial Differential Equations, and the Methusalem programme of the Ghent University Special Research Fund (BOF) (Grant number 01M01021). Michael Ruzhansky is also supported by EPSRC grant EP/R003025/2. M. Chatzakou is a postdoctoral fellow of the Research Foundation – Flanders (FWO) under the postdoctoral grant No 12B1223N.

\bibliographystyle{elsarticle-num}

\bibliography{main}

\end{document}